\documentclass[10pt,twocolumn,letterpaper]{article}

\usepackage{cvpr}
\usepackage{times}
\usepackage{epsfig}
\usepackage{graphicx}
\usepackage{amsmath}
\usepackage{amssymb}

\usepackage{comment}
\usepackage[british,UKenglish,USenglish,english,american]{babel}
\usepackage{multirow}

\newcommand{\tabincell}[2]{\begin{tabular}{@{}#1@{}}#2\end{tabular}}

\usepackage[pagebackref=true,breaklinks=true,letterpaper=true,colorlinks,bookmarks=false]{hyperref}

\cvprfinalcopy 

\ifcvprfinal\fi
\begin{document}

\title{Fully Convolutional Instance-aware Semantic Segmentation}

\author{Yi Li$^{1,2}$\thanks{Equal contribution. This work is done when Yi Li and Haozhi Qi are interns at Microsoft Research.} \qquad Haozhi Qi$^{2*}$ \qquad Jifeng Dai$^{2}$ \qquad Xiangyang Ji$^1$ \qquad Yichen Wei$^2$\\
$^1$Tsinghua University \qquad \qquad \qquad $^2$Microsoft Research Asia \quad\quad\quad\quad\quad\quad\\
{\tt\small \{liyi14,xyji\}@tsinghua.edu.cn, \{v-haoq,jifdai,yichenw\}@microsoft.com}
}

\maketitle

\begin{abstract}
We present the first fully convolutional end-to-end solution for instance-aware semantic segmentation task. It inherits all the merits of FCNs for semantic segmentation~\cite{long2015fully} and instance mask proposal~\cite{dai2016instance}. It detects and segments the object instances jointly and simultanoulsy. By the introduction of position-senstive inside/outside score maps, the underlying convolutional representation is fully shared between the two sub-tasks, as well as between all regions of interest. The proposed network is highly integrated and achieves state-of-the-art performance in both accuracy and efficiency. It wins the COCO 2016 segmentation competition by a large margin. Code would be released at \url{https://github.com/daijifeng001/TA-FCN}.

\end{abstract}

\section{Introduction}
\label{sec.intro}
\begin{figure*}
	\centering
	\includegraphics[width=1.0\linewidth]{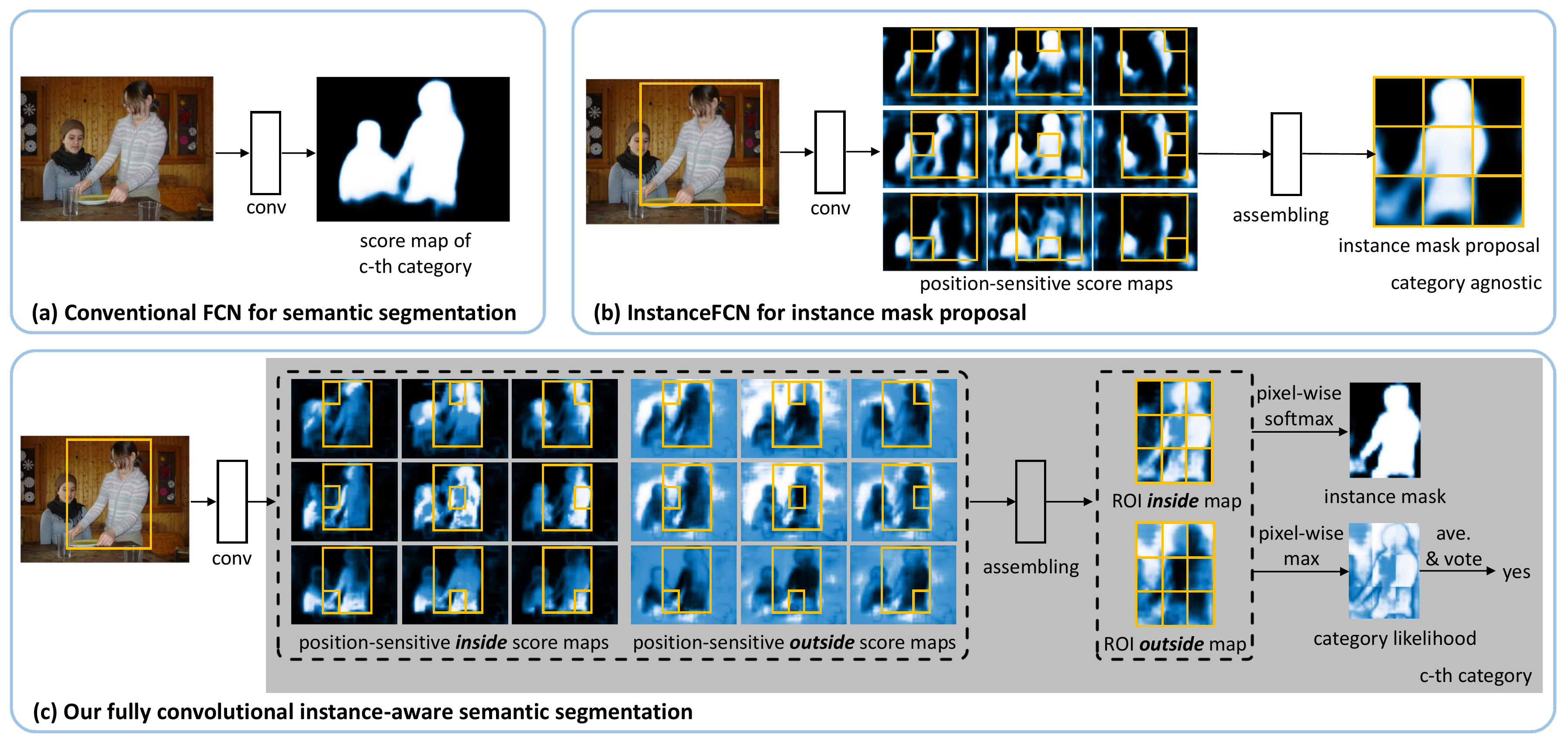}
	\caption{Illustration of our idea. (a) Conventional fully convolutional network (FCN)~\cite{long2015fully} for semantic segmentation. A single score map is used for each category, which is unaware of individual object instances. (b) InstanceFCN~\cite{dai2016instance} for instance segment proposal, where $3\times 3$ position-sensitive score maps are used to encode relative position information. A downstream network is used for segment proposal classification. (c) Our fully convolutional instance-aware semantic segmentation method (FCIS), where position-sensitive inside/outside score maps are used to perform object segmentation and detection jointly and simultanously.}
	\label{fig:instance_sensitive_score_map}
\end{figure*}

Fully convolutional networks (FCNs)~\cite{long2015fully} have recently dominated the field of semantic image segmentation. An FCN takes an input image of arbitrary size, applies a series of convolutional layers, and produces per-pixel likelihood score maps for all semantic categories, as illustrated in Figure~\ref{fig:instance_sensitive_score_map}(a). Thanks to the simplicity, efficiency, and the local weight sharing property of convolution, FCNs provide an accurate, fast, and end-to-end solution for semantic segmentation.

\begin{figure*}
	\centering
	\includegraphics[width=0.9\linewidth]{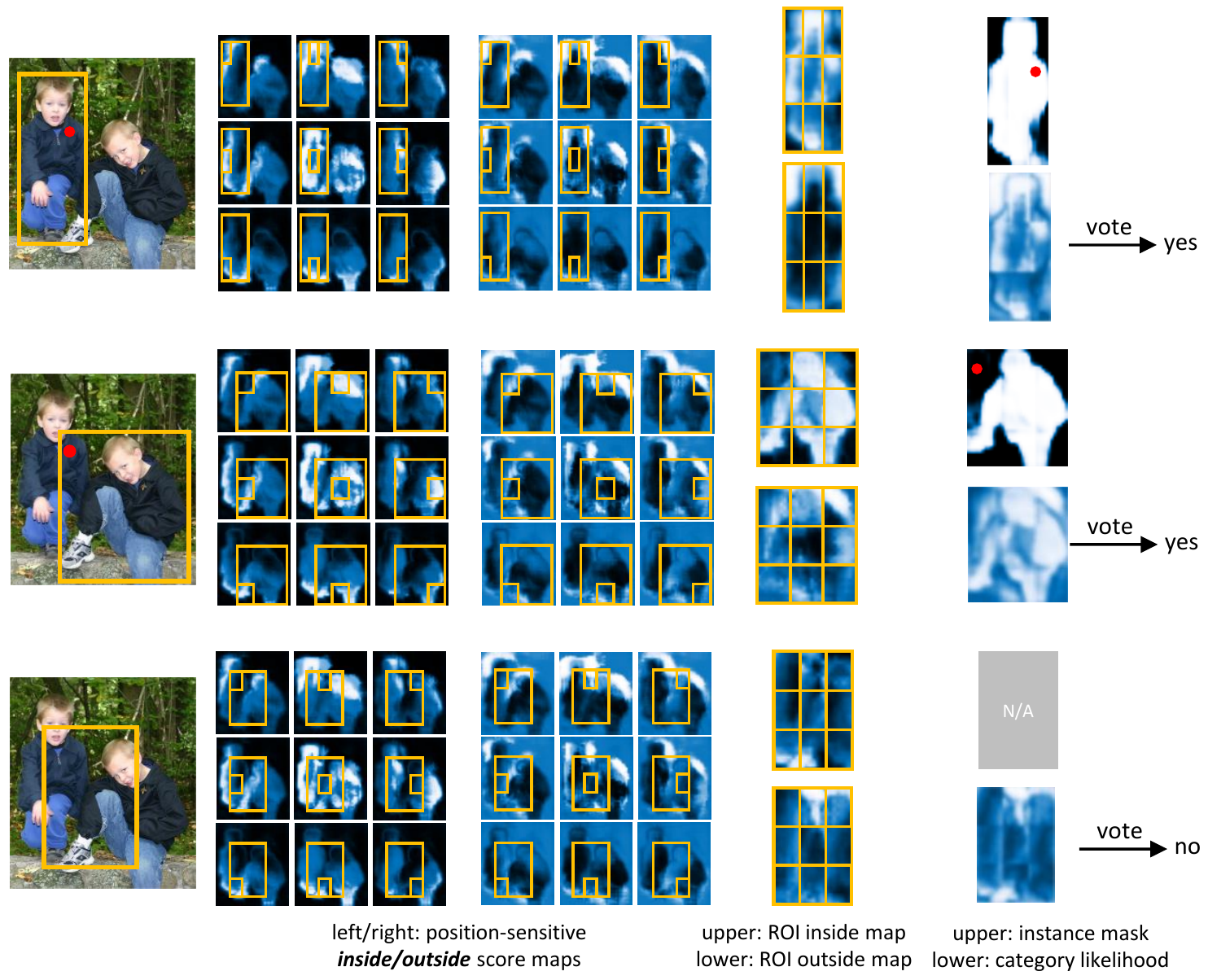}
	\caption{Instance segmentation and classification results (of ``person" category) of different ROIs. The score maps are shared by different ROIs and both sub-tasks. The red dot indicates one pixel having different semantics in different ROIs.}
	\label{fig:instance_segmentaion_example}
\end{figure*}

However, conventional FCNs do not work for the instance-aware semantic segmentation task, which requires the detection and segmentation of individual object instances. The limitation is inherent. Because convolution is translation invariant, the same image pixel receives the same responses (thus classification scores) irrespective to its relative position in the context. However, instance-aware semantic segmentation needs to operate on region level, and the same pixel can have different semantics in different regions. This behavior cannot be modeled by a single FCN on the whole image. The problem is exemplified in Figure~\ref{fig:instance_segmentaion_example}.

Certain translation-variant property is required to solve the problem. In a prevalent family of instance-aware semantic segmentation approaches~\cite{dai2015convolutional,hariharan2015hyper,dai2016mnc}, it is achieved by adopting different types of sub-networks in three stages: 1) an FCN is applied on the whole image to generate intermediate and shared feature maps; 2) from the shared feature maps, a pooling layer warps each region of interest (ROI) into fixed-size per-ROI feature maps~\cite{he2014spatial,girshick2015fast}; 3) one or more fully-connected (fc) layer(s) in the last network convert the per-ROI feature maps to per-ROI masks. Note that the translation-variant property is introduced in the fc layer(s) in the last step.

Such methods have several drawbacks. First, the ROI pooling step losses spatial details due to feature warping and resizing, which however, is necessary to obtain a fixed-size representation (\eg, $14 \times 14$ in~\cite{dai2016mnc}) for fc layers. Such distortion and fixed-size representation degrades the segmentation accuracy, especially for large objects. Second, the fc layers over-parametrize the task, without using regularization of local weight sharing. For example, the last fc layer has high dimensional 784-way output to estimate a $28 \times 28$ mask. Last, the per-ROI network computation in the last step is not shared among ROIs. As observed empirically, a considerably complex sub-network in the last step is necessary to obtain good accuracy~\cite{ren2015object,dai2016rfcn}. It is therefore slow for a large number of ROIs (typically hundreds or thousands of region proposals). For example, in the MNC method~\cite{dai2016mnc}, which won the 1st place in COCO segmentation challenge 2015~\cite{lin2014coco}, 10 layers in the ResNet-101 model~\cite{he2016deep} are kept in the per-ROI sub-network. The approach takes $1.4$ seconds per image, where more than $80\%$ of the time is spent on the last per-ROI step. These drawbacks motivate us to ask the question that, \emph{can we exploit the merits of FCNs for end-to-end instance-aware semantic segmentation?}

Recently, a fully convolutional approach has been proposed for instance mask proposal generation~\cite{dai2016instance}. It extends the translation invariant score maps in conventional FCNs to \emph{position-sensitive} score maps, which are somewhat translation-variant. This is illustrated in Figure~\ref{fig:instance_sensitive_score_map}(b). The approach is only used for mask proposal generation and presents several drawbacks. It is blind to semantic categories and requires a downstream network for detection. The object segmentation and detection sub-tasks are separated and the solution is not end-to-end. It operates on square, fixed-size sliding windows ($224 \times 224$ pixels) and adopts a time-consuming image pyramid scanning to find instances at different scales.

In this work, we propose the first end-to-end fully convolutional approach for instance-aware semantic segmentation. Dubbed FCIS, it extends the approach in~\cite{dai2016instance}. The underlying convolutional representation and the score maps are fully shared for the object segmentation and detection sub-tasks, via a novel joint formulation with no extra parameters. The network structure is highly integrated and efficient. The per-ROI computation is simple, fast, and does not involve any warping or resizing operations. The approach is briefly illustrated in Figure~\ref{fig:instance_sensitive_score_map}(c). It operates on box proposals instead of sliding windows, enjoying the recent advances in object detection~\cite{ren2015faster}.

Extensive experiments verify that the proposed approach is state-of-the-art in both accuracy and efficiency. It achieves significantly higher accuracy than the previous challenge winning method MNC~\cite{dai2016mnc} on the large-scale COCO dataset~\cite{lin2014coco}. It wins the 1st place in COCO 2016 segmentation competition, outperforming the 2nd place entry by $12\%$ in accuracy relatively. It is fast. The inference in COCO competition takes 0.24 seconds per image using ResNet-101 model~\cite{he2016deep} (Nvidia K40), which is $6\times$ faster than MNC~\cite{dai2016mnc}. Code would be released at \url{https://github.com/daijifeng001/TA-FCN}.

\section{Our Approach}

\subsection{Position-sensitive Score Map Parameterization}
\label{sec.position_sensitive_score_review}
In FCNs~\cite{long2015fully}, a classifier is trained to predict each pixel's likelihood score of ``\emph{the pixel belongs to some object category}''. It is translation invariant and unaware of individual object instances. For example, the same pixel can be foreground on one object but background on another (adjacent) object. A single score map per-category is insufficient to distinguish these two cases.

To introduce translation-variant property, a fully convolutional solution is firstly proposed in~\cite{dai2016instance} for instance mask proposal. It uses $k^2$ position-sensitive score maps that correspond to $k\times k$ evenly partitioned cells of objects. This is illustrated in Figure~\ref{fig:instance_sensitive_score_map}(b) ($k=3$). Each score map has the same spatial extent of the original image (in a lower resolution, \eg., $16\times$ smaller). Each score represents the likelihood of ``\emph{the pixel belongs to some object instance at a relative position}''. For example, the first map is for ``at top left position'' in Figure~\ref{fig:instance_sensitive_score_map}(b).

During training and inference, for a fixed-size square sliding window ($224\times224$ pixels), its pixel-wise foreground likelihood map is produced by assembling (copy-paste) its $k\times k$ cells from the corresponding score maps. In this way, a pixel can have different scores in different instances as long as the pixel is at different relative positions in the instances.

As shown in~\cite{dai2016instance}, the approach is state-of-the-art for the object mask proposal task. However, it is also limited by the task. Only a fixed-size square sliding window is used. The network is applied on multi-scale images to find object instances of different sizes. The approach is blind to the object categories. Only a separate ``objectness'' classification sub-network is used to categorize the window as object or background. For the instance-aware semantic segmentation task, a separate downstream network is used to further classify the mask proposals into object categories~\cite{dai2016instance}.

\subsection{Joint Mask Prediction and Classification}
\label{sec.joint_formulation}
For the instance-aware semantic segmentation task, not only ~\cite{dai2016instance}, but also many other state-of-the-art approaches, such as SDS~\cite{hariharan2014simultaneous}, Hypercolumn~\cite{hariharan2015hyper}, CFM~\cite{dai2015convolutional}, MNC~\cite{dai2016mnc}, and MultiPathNet \cite{zagoruyko2016multipath}, share a similar structure: two sub-networks are used for object segmentation and detection sub-tasks, \emph{separately and sequentially}.

Apparently, the design choices in such a setting, \eg, the two networks' structure, parameters and execution order, are kind of arbitrary. They can be easily made for convenience other than for fundamental considerations. We conjecture that the separated sub-network design may not fully exploit the tight correlation between the two tasks.

We enhance the ``position-sensitive score map'' idea to perform the object segmentation and detection sub-tasks \emph{jointly and simultaneously}. The same set of score maps are shared for the two sub-tasks, as well as the underlying convolutional representation. Our approach brings no extra parameters and eliminates non essential design choices. We believe it can better exploit the strong correlation between the two sub-tasks.

Our approach is illustrated in Figure~\ref{fig:instance_sensitive_score_map}(c) and Figure~\ref{fig:instance_segmentaion_example}. Given a region-of-interest (ROI), its pixel-wise score maps are produced by the assembling operation within the ROI. For each pixel in a ROI, there are two tasks: 1) detection: whether it belongs to an object bounding box at a relative position (detection+) or not (detection-); 2) segmentation: whether it is inside an object instance's boundary (segmentation+) or not (segmentation-). A simple solution is to train two classifiers, separately. That's exactly our baseline \emph{FCIS (separate score maps)} in Table~\ref{tab:ablation_voc}. In this case, the two classifiers are two $1\times 1$ conv layers, each using just one task's supervision.

Our joint formulation fuses the two answers into two scores: inside and outside. There are three cases: 1) high inside score and low outside score: detection+, segmentation+; 2) low inside score and high outside score: detection+, segmentation-; 3) both scores are low: detection-, segmentation-. The two scores answer the two questions jointly via softmax and max operations. For detection, we use max to differentiate cases 1)-2) (detection+) from case 3) (detection-). The detection score of the whole ROI is then obtained via average pooling over all pixels' likelihoods (followed by a softmax operator across all the categories). For segmentation, we use softmax to differentiate cases 1) (segmentation+) from 2) (segmentation-), at each pixel. The foreground mask (in probabilities) of the ROI is the union of the per-pixel segmentation scores (for each category). Similarly, the two sets of scores are from two $1\times 1$ conv layer. The inside/outside classifiers are trained jointly as they receive the back-propagated gradients from both segmentation and detection losses. 

The approach has many desirable properties. All the per-ROI components (as in Figure~\ref{fig:instance_sensitive_score_map}(c)) do not have free parameters. The score maps are produced by a single FCN, without involving any feature warping, resizing or fc layers. All the features and score maps respect the aspect ratio of the original image. The local weight sharing property of FCNs is preserved and serves as a regularization mechanism. All per-ROI computation is simple ($k^2$ cell division, score map copying, softmax, max, average pooling) and fast, giving rise to a negligible per-ROI computation cost.

\begin{figure*}
	\centering
	\includegraphics[width=0.85\linewidth]{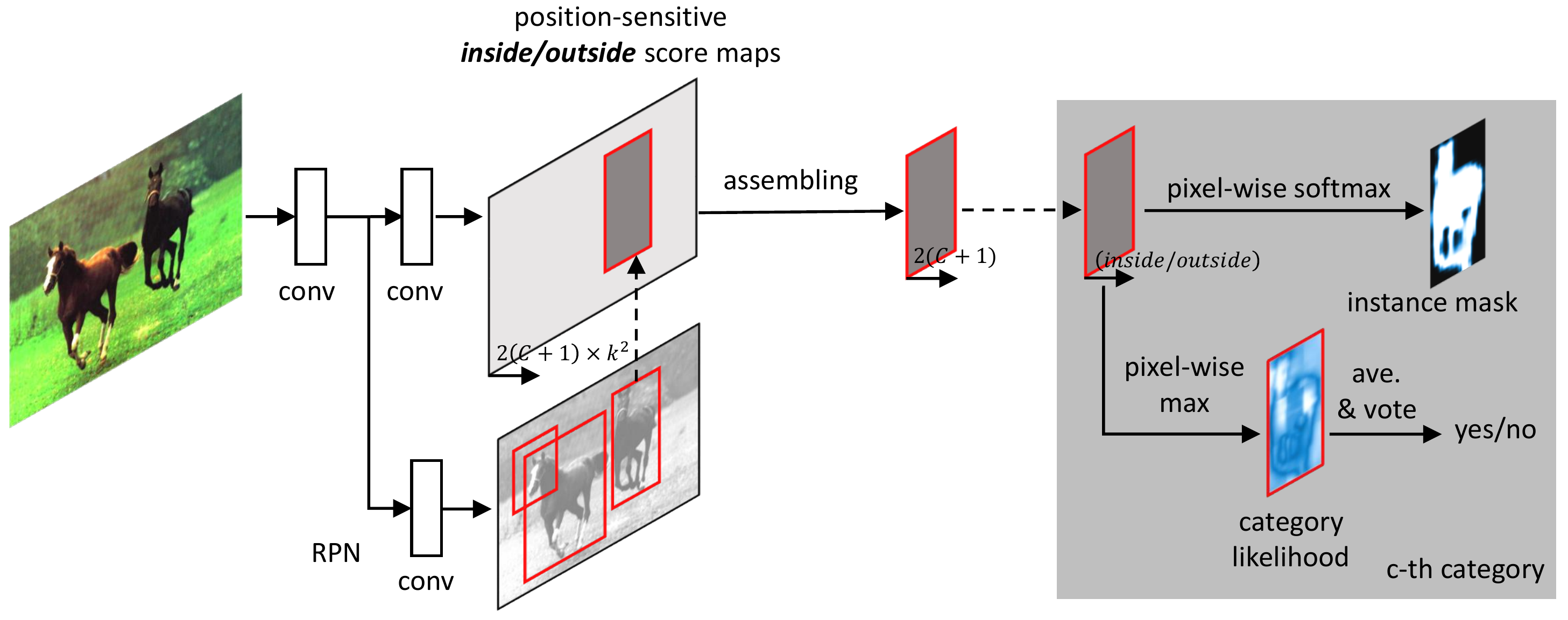}
	\caption{Overall architecture of FCIS. A region proposal network (RPN)~\cite{ren2015faster} shares the convolutional feature maps with FCIS. The proposed region-of-interests (ROIs) are applied on the score maps for joint object segmentation and detection. The learnable weight layers are fully convolutional and computed on the whole image. The per-ROI computation cost is negligible.}
	\label{fig:network_architecture}
\end{figure*}

\subsection{An End-to-End Solution}
\label{sec.end_to_end}
Figure~\ref{fig:network_architecture} shows the architecture of our end-to-end solution. While any convolutional network architecture can be used~\cite{simonyan2015very,szegedy2015going}, in this work we adopt the ResNet model~\cite{he2016deep}. The last fully-connected layer for $1000-$way classification is discarded. Only the previous convolutional layers are retained. The resulting feature maps have $2048$ channels. On top of it, a $1\times 1$ convolutional layer is added to reduce the dimension to $1024$.

In the original ResNet, the effective feature stride (the decrease in feature map resolution) at the top of the network is 32. This is too coarse for instance-aware semantic segmentation. To reduce the feature stride and maintain the field of view, the ``hole algorithm"~\cite{chen2015semantic,long2015fully} (\emph{Algorithme \`{a} trous} \cite{mallat1999wavelet}) is applied. The stride in the first block of conv5 convolutional layers is decreased from 2 to 1. The effective feature stride is thus reduced to 16. To maintain the field of view, the ``hole algorithm" is applied on all the convolutional layers of conv5 by setting the dilation as 2.

We use region proposal network (RPN)~\cite{ren2015faster} to generate ROIs. For fair comparison with the MNC method~\cite{dai2016mnc}, it is added on top of the conv4 layers in the same way. Note that RPN is also fully convolutional.

From the conv5 feature maps, $2k^2\times (C+1)$ score maps are produced ($C$ object categories, one background category, two sets of $k^2$ score maps per category, $k=7$ by default in experiments) using a $1\times 1$ convolutional layer. Over the score maps, each ROI is projected into a $16\times$ smaller region. Its segmentation probability maps and classification scores over all the categories are computed as described in Section~\ref{sec.joint_formulation}.

Following the modern object detection systems, bounding box (bbox) regression~\cite{girshick2014rich,girshick2015fast} is used to refine the initial input ROIs. A sibling $1\times 1$ convolutional layer with $4k^2$ channels is added on the conv5 feature maps to estimate the bounding box shift in location and size.

Below we discuss more details in inference and training.

\textbf{Inference}
For an input image, $300$ ROIs with highest scores are generated from RPN. They pass through the bbox regression branch and give rise to another $300$ ROIs. For each ROI, we get its classification scores and foreground mask (in probability) for all categories. Figure~\ref{fig:instance_segmentaion_example} shows an example. Non-maximum suppression (NMS) with an intersection-over-union (IoU) threshold $0.3$ is used to filter out highly overlapping ROIs. The remaining ROIs are classified as the categories with highest classification scores. Their foreground masks are obtained by mask voting~\cite{dai2016mnc} as follows. For an ROI under consideration, we find all the ROIs (from the $600$) with IoU scores higher than 0.5. Their foreground masks of the category are averaged on a per-pixel basis, weighted by their classification scores. The averaged mask is binarized as the output.

\textbf{Training}
An ROI is positive if its box IoU with respect to the nearest ground truth object is larger than $0.5$, otherwise it is negative. Each ROI has three loss terms in equal weights: a softmax detection loss over $C+1$ categories, a softmax segmentation loss~\footnote{The term sums per-pixel losses over the ROI and normalizes the sum by the ROI's size.} \emph{over the foreground mask of the ground-truth category only}, and a bbox regression loss as in~\cite{girshick2015fast}. The latter two loss terms are effective only on the positive ROIs.

During training, the model is initialized from the pre-trained model on ImageNet classification~\cite{he2016deep}. Layers absent in the pre-trained model are randomly initialized. The training images are resized to have a shorter side of 600 pixels. We use SGD optimization. We train the model using 8 GPUs, each holding one image mini batch, giving rise to an effective batch size $\times 8$. For experiments on PASCAL VOC~\cite{everingham2010pascal}, 30k iterations are performed, where the learning rates are $10^{-3}$ and $10^{-4}$ in the first 20k and the last 10k iterations respectively. The iteration number is $\times8$ for experiments on COCO~\cite{lin2014coco}.

As the per-ROI computation is negligible, the training benefits from inspecting more ROIs at small training cost. Specifically, we apply online hard example mining (OHEM)~\cite{shrivastava2016training}. In each mini batch, forward propagation is performed on all the 300 proposed ROIs on one image. Among them, 128 ROIs with the highest losses are selected to back-propagate their error gradients.

For the RPN proposals, 9 anchors (3 scales $\times$ 3 aspect ratios) are used by default. 3 additional anchors at a finer scale are used for experiments on the COCO dataset~\cite{lin2014coco}. To enable feature sharing between FCIS and RPN, joint training is performed~\cite{dai2016mnc,ren2016faster}.

\section{Related Work}

\textbf{Semantic Image Segmentation} The task is to assign every pixel in the image a semantic category label. It does not distinguish object instances. Recently, this field has been dominated by a prevalent family of approaches based on FCNs~\cite{long2015fully}. The FCNs are extended with global context~\cite{liu2016parsenet}, multi-scale feature fusion~\cite{chen2016attention}, and deconvolution~\cite{noh2015deconvolution}. Recent works in~\cite{chen2015semantic,zheng2015conditional,schwing2015fully,lin2016piecewise} integrated FCNs with conditional random fields (CRFs). The expensive CRFs are replaced by more efficient domain transform in~\cite{chen2016taskspecific}. As the per-pixel category labeling is expensive, the supervision signals in FCNs have been relaxed to boxes~\cite{dai2015boxsup}, scribbles~\cite{lin2016ScribbleSup}, or weakly supervised image class labels~\cite{hong2015decoupled,hong2016transferrable}.

\textbf{Object Segment Proposal} The task is to generate category-agnostic object segments. Traditional approaches, \eg, MCG~\cite{arbelaez2014mcg} and Selective Search~\cite{sande2011selective}, use low level image features. Recently, the task is achieved by deep learning approaches, such as DeepMask~\cite{pinheiro2015learning} and SharpMask~\cite{pinheiro2016refine}. Recently, a fully convolutional approach is proposed in~\cite{dai2016instance}, which inspires this work.

\textbf{Instance-aware Semantic Segmentation} The task requires both classification and segmentation of object instances. Typically, the two sub-tasks are accomplished separately. Usually, the segmentation task relies on a segment proposal method and the classification task is built on the region-based methods~\cite{girshick2014rich,girshick2015fast,ren2015faster}. This paradigm includes most state-of-the-art approaches, such as SDS~\cite{hariharan2014simultaneous}, Hypercolumn~\cite{hariharan2015hyper}, CFM~\cite{dai2015convolutional}, MNC~\cite{dai2016mnc}, MultiPathNet~\cite{zagoruyko2016multipath}, and iterative approach~\cite{li2016iterative}. Such approaches have certain drawbacks, as discussed in Section~\ref{sec.intro} and Section~\ref{sec.joint_formulation}. In this work, we propose a fully convolutional approach with an integrated joint formulation for the two sub-tasks.

There are some endeavors~\cite{liang2015proposal,liu2016multiscale} trying to extend FCNs for instance-aware semantic segmentation, by grouping/clustering the FCN's output. However, all these methods rely on complex hand-crafted post processing, and are not end-to-end. The performance is also not satisfactory.

\textbf{FCNs for Object Detection}
The idea of ``position sensitive score maps'' in~\cite{dai2016instance} is adapted in R-FCN~\cite{dai2016rfcn}, resulting in a fully convolutional approach for object detection. The score maps are re-purposed from foreground-background segmentation likelihood to object category likelihood. R-FCN~\cite{dai2016rfcn} only performs object classification. It is unaware of the instance segmentation task. Yet, it can be combined with~\cite{dai2016instance} for instance-aware semantic segmentation task, in a straightforward manner. This is investigated in our experiments (Section~\ref{sec.exp}).

\section{Experiments}

\subsection{Ablation Study on PASCAL VOC}
\label{sec.exp}

Ablation experiments are performed to study the proposed FCIS method on the PASCAL VOC dataset~\cite{everingham2010pascal}. Following the protocol in \cite{hariharan2014simultaneous,dai2015convolutional,hariharan2015hyper,dai2016mnc}, model training is performed on the VOC 2012 train set, and evaluation is performed on the VOC 2012 validation set, with the additional instance mask annotations from \cite{hariharan2011semantic}. Accuracy is evaluated by mean average precision, mAP$^r$~\cite{hariharan2014simultaneous}, at mask-level IoU (intersection-over-union) thresholds at $0.5$ and $0.7$.

The proposed \textbf{FCIS} approach is compared with alternative (almost) fully convolutional baseline methods, as well as variants of FCIS with different design choices. For fair comparison, ImageNet~\cite{deng2009imagenet} pre-trained ResNet-101 model~\cite{he2016deep} is used for all the methods. OHEM is not applied.

\textbf{na\"{\i}ve MNC}. This baseline is similar to MNC~\cite{dai2016mnc} except that all convolutional layers of ResNet-101 are applied on the whole image to obtain feature maps, followed by ROI pooling on top of the last block of conv5 layers. A 784-way fc layer is applied on the ROI pooled features for mask prediction (of resolution $28 \times 28$), together with a 21-way fc layer for classification. The \`{a} trous trick is also applied for fair comparison. It is almost fully convolutional, with only single layer fc sub-networks in per-ROI computation.

\textbf{InstFCN + R-FCN}. The class-agnostic mask proposals are firstly generated by InstFCN~\cite{dai2016instance}, and then classified by R-FCN~\cite{dai2016rfcn}. It is a straightforward combination of InstFCN and R-FCN. The two FCNs are separately trained and applied for mask prediction and classification, respectively.

\textbf{FCIS (translation invariant)}. To verify the importance of the translation-variant property introduced by the position sensitive score maps, this baseline sets $k=1$ in the FCIS method to make it translation invariant.

\textbf{FCIS (separate score maps)}. To validate the joint formulation for mask prediction and classification, this baseline uses the two sets of score maps separately for the two sub-tasks. The first set of $k^2$ score maps are only for segmentation, in the similar way as in~\cite{dai2016instance}. The second set is only for classification, in the same way as in R-FCN~\cite{dai2016rfcn}. Therefore, the preceding convolutional classifiers for the two sets of score maps are not related, while the shallower convolutional feature maps are still shared.

\setlength{\tabcolsep}{5pt}
\renewcommand{\arraystretch}{1.5}
\begin{table}
\begin{center}
\small
\begin{tabular}{l|c|c}
\hline
\footnotesize method & \scriptsize mAP$^r$@0.5 (\%) & \scriptsize mAP$^r$@0.7 (\%) \\
\hline
\hline
\footnotesize na\"{\i}ve MNC                      				 & 59.1 & 36.0 \\
\hline
\footnotesize InstFCN + R-FCN                                    & 62.7 & 41.5 \\
\hline
\footnotesize FCIS \scriptsize (translation invariant)    		 & 52.5 & 38.5 \\
\hline
\footnotesize FCIS \scriptsize (separate score maps)             & 63.9 & 49.7 \\
\hline
\footnotesize FCIS                                           & \underline{65.7} & \underline{52.1} \\
\hline
\end{tabular}
\end{center}
\caption{Ablation study of (almost) fully convolutional methods on PASCAL VOC 2012 validation set.}
\label{tab:ablation_voc}
\end{table}

Table \ref{tab:ablation_voc} shows the results. The  mAP$^r$ scores of the na\"{\i}ve MNC baseline are 59.1\% and 36.0\% at IoU thresholds of 0.5 and 0.7 respectively. They are 5.5\% and 12.9\% lower than those of the original MNC~\cite{dai2016mnc}, which keeps 10 layers in ResNet-101 in the per-ROI sub-networks. This verifies the importance of respecting the translation-variant property for instance-aware semantic segmentation.

The result of ``InstFCN + R-FCN" is reasonably good, but is still inferior than that of FCIS. The inference speed is also slow (1.27 seconds per image on a Nvidia K40 GPU).

The proposed FCIS method achieves the best result. This verifies the effectiveness of our end-to-end solution. Its degenerated version ``FCIS (translation invariant)" is much worse, indicating the position sensitive score map parameterization is vital. Its degenerated version ``FCIS (separate score maps)" is also worse, indicating that the joint formulation is effective.

\subsection{Experiments on COCO}
Following the COCO~\cite{lin2014coco} experiment guideline, training is performed on the 80k+40k trainval images, and results are reported on the test-dev set. We evaluate the performance using the standard COCO evaluation metric, mAP$^r$@[0.5:0.95], as well as the traditional mAP$^r$@0.5 metric.

\paragraph{Comparison with MNC}

\setlength{\tabcolsep}{2pt}
\renewcommand{\arraystretch}{1.2}
\begin{table*}
\begin{center}
\small
\begin{tabular}{l|c|c|c|c|c|c|c|c}
\hline
\footnotesize method & \renewcommand{\arraystretch}{1.1} \scriptsize \tabincell{c}{sampling strategy \\ in training} & \scriptsize train time/img & \scriptsize test time/img & \scriptsize mAP$^r$@[0.5:0.95] (\%) & \scriptsize mAP$^r$@0.5 (\%) & \renewcommand{\arraystretch}{1.1} \scriptsize \tabincell{c}{mAP$^r$@[0.5:0.95] (\%) \\ (small)} & \renewcommand{\arraystretch}{1.1} \scriptsize \tabincell{c}{mAP$^r$@[0.5:0.95] (\%) \\ (mid)} & \renewcommand{\arraystretch}{1.1} \scriptsize \tabincell{c}{mAP$^r$@[0.5:0.95] (\%) \\ (large)}\\
\hline
\hline
\footnotesize MNC                & \footnotesize random & 2.05s & 1.37s & 24.6 & 44.3 & 4.7 & 25.9 & 43.6 \\
\footnotesize \textbf{FCIS}   & \footnotesize random & 0.53s & 0.24s & \underline{28.8} & \underline{48.7} & \underline{6.8} & \underline{30.8} & \underline{49.5} \\
\hline
\footnotesize MNC                & \footnotesize OHEM & 3.22s & 1.37s & N/A & N/A & N/A & N/A & N/A \\
\footnotesize \textbf{FCIS}   & \footnotesize OHEM & 0.54s & 0.24s & \textbf{29.2} & \textbf{49.5} & \textbf{7.1} & \textbf{31.3} & \textbf{50.0} \\
\hline
\end{tabular}
\end{center}
\caption{Comparison with MNC~\cite{dai2016mnc} on COCO test-dev set, using ResNet-101 model. Timing is evaluated on a Nvidia K40 GPU.}
\label{tab:comparison_mnc_coco}
\end{table*}

We compare the proposed FCIS method with MNC~\cite{dai2016mnc}, the 1st place entry in COCO segmentation challenge 2015. Both methods perform mask prediction and classification in ROIs, and share similar training/inference procedures. For fair comparison, we keep their common implementation details the same.

Table \ref{tab:comparison_mnc_coco} presents the results using ResNet-101 model. When OHEM is not used, FCIS achieves an mAP$^r$@[0.5:0.95] score of 28.8\% on COCO test-dev set, which is 4.2\% absolutely (17\% relatively) higher than that of MNC. According to the COCO standard split of object sizes, the accuracy improvement is more significant for larger objects, indicating that FCIS can capture the detailed spatial information better. FCIS is also much faster than MNC. In inference, FCIS spends 0.24 seconds per image on a Nvidia K40 GPU (0.19 seconds for network forward, and 0.05 seconds for mask voting), which is $\sim6 \times$ faster than MNC. FCIS is also $\sim4 \times$ faster in training. In addition, FCIS easily benefits from OHEM due to its almost free per-ROI cost, achieving an mAP$^r$@[0.5:0.95] score of 29.2\%. Meanwhile, OHEM is unaffordable for MNC, because considerable computational overhead would be added during training.

\paragraph{Networks of Different Depths}

\setlength{\tabcolsep}{4pt}
\renewcommand{\arraystretch}{1.2}
\begin{table}
\begin{center}
\small
\begin{tabular}{l|c|c|c}
\hline
\scriptsize network architecture & \scriptsize mAP$^r$@[0.5:0.95] (\%) & \scriptsize mAP$^r$@0.5 (\%) & \scriptsize test time/img\\
\hline
\hline
\footnotesize ResNet-50    & 27.1 & 46.7 & 0.16s \\
\footnotesize ResNet-101   & 29.2 & 49.5 & 0.24s \\
\footnotesize ResNet-152   & 29.5 & 49.8 & 0.27s \\
\hline
\end{tabular}
\end{center}
\caption{Results of using networks of different depths in FCIS.}
\label{tab:depth_coco}
\end{table}

Table \ref{tab:depth_coco} presents the results of using ResNet of different depths in FCIS method. The accuracy is improved when the network depth is increased from 50 to 101, and gets saturated when the depth reaches 152.

\paragraph{COCO Segmentation Challenge 2016 Entry}

\setlength{\tabcolsep}{5pt}
\renewcommand{\arraystretch}{1.2}
\begin{table}
\begin{center}
\small
\begin{tabular}{l|c|c}
\hline
 & \scriptsize mAP$^r$@[0.5:0.95] (\%) & \scriptsize mAP$^r$@0.5 (\%) \\
\hline
\hline
\footnotesize FAIRCNN (2015)                  & 25.0 & 45.6 \\
\footnotesize MNC+++ (2015)                   & 28.4 & 51.6 \\
\footnotesize G-RMI (2016)                    & 33.8 & 56.9 \\
\hline
\footnotesize FCIS baseline                   & 29.2 & 49.5 \\
\footnotesize +multi-scale testing            & 32.0 & 51.9 \\
\footnotesize +horizontal flip                & 32.7 & 52.7 \\
\footnotesize +multi-scale training           & 33.6 & 54.5 \\
\footnotesize +ensemble                       & \textbf{37.6} & \textbf{59.9} \\
\hline
\end{tabular}
\end{center}
\caption{Instance-aware semantic segmentation results of different entries for the COCO segmentation challenge (2015 and 2016) on COCO test-dev set.}
\label{tab:challenge_coco}
\end{table}

Based on the FCIS method, we participated in COCO segmentation challenge 2016 and won the 1st place.

Table \ref{tab:challenge_coco} presents the results of our entry and other entries in COCO segmentation challenge 2015 and 2016. Our entry is based on FCIS, with some simple bells and whistles.

\emph{FCIS Baseline}. The baseline FCIS method achieves a competitive mAP$^r$@[0.5:0.95] score of 29.2\%, which is already higher than MNC+++~\cite{dai2016mnc}, the winning entry in 2015. 

\emph{Multi-scale testing}. Following~\cite{he2014spatial,he2016deep}, the position-sensitive score maps are computed on a pyramid of testing images, where the shorter sides are of $\{480, 576, 688, 864, 1200, 1400\}$ pixels. For each ROI, we obtain its result from the scale where the ROI has a number of pixels closest to $224\times224$. Note that RPN proposals are still computed from a single scale (shorter side 600). Multi-scale testing improves the accuracy by 2.8\%. 

\emph{Horizontal flip}. Similar to \cite{zagoruyko2016multipath}, the FCIS method is applied on the original and the flipped images, and the results in the corresponding ROIs are averaged. This helps increase the accuracy by 0.7\%. 

\emph{Multi-scale training}. We further apply multi-scale training at the same scales as in multi-scale inference. For the finer scales, a random $600\times600$ image patch is cropped for training due to memory issues, as in \cite{liu2016ssd}. This increases the accuracy by 0.9\%. 

\begin{figure*}
\centering
\includegraphics[width=0.9\linewidth]{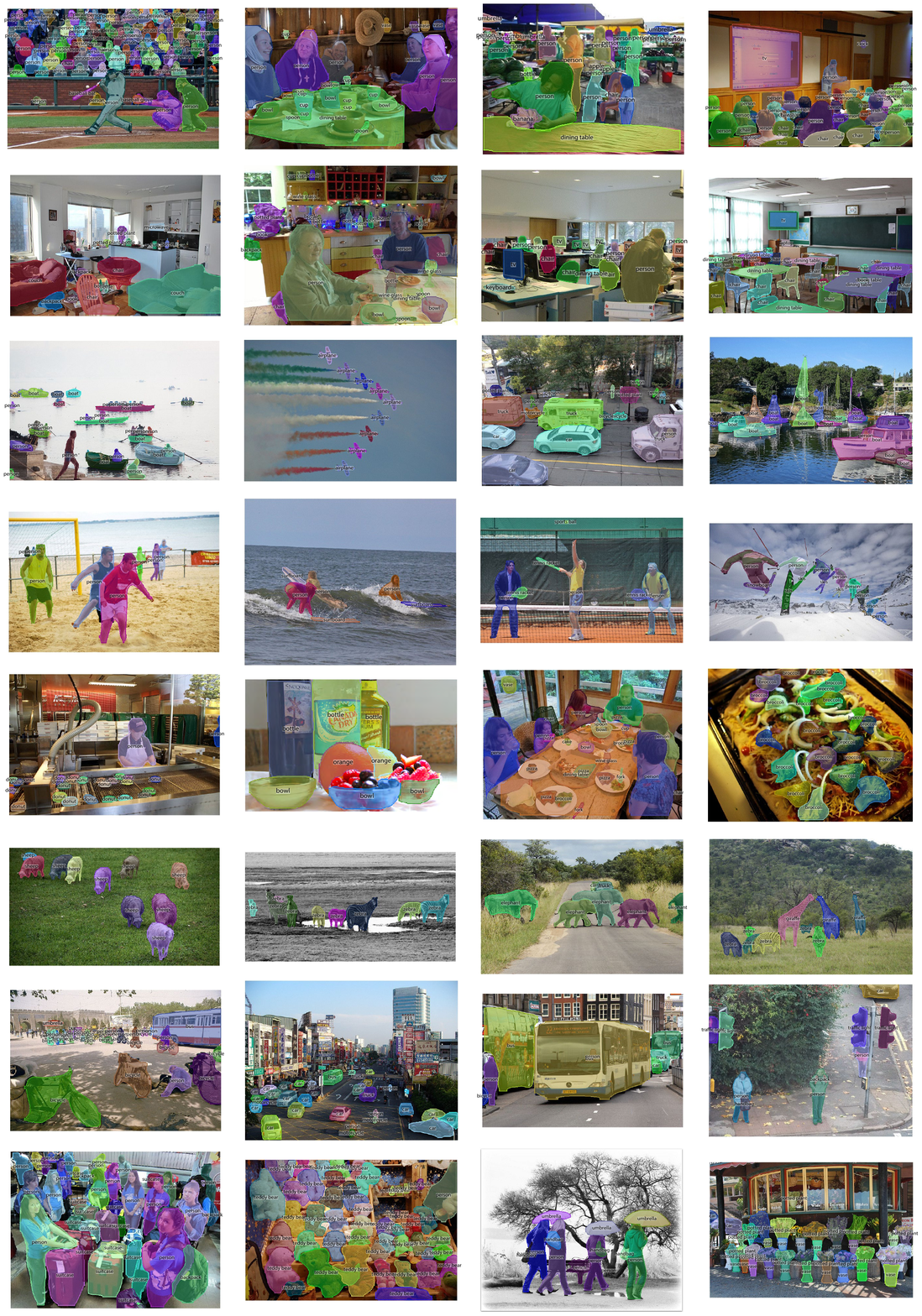}
\caption{Example instance-aware semantic segmentation results of the proposed FCIS method on COCO test set. Check \url{https://github.com/daijifeng001/TA-FCN} for example results on the first 5k images on COCO test set.}
\label{fig:visualization_coco}
\end{figure*}

\emph{Ensemble}. Following \cite{he2016deep}, region proposals are generated using an ensemble, and the union of the proposals are processed by an ensemble for mask prediction and classification. We utilize an ensemble of 6 networks. The final result is 37.6\%, which is 3.8\% (11\% relatively) higher than G-RMI, the 2nd place entry in 2016, and 9.2\% (32\% relatively) higher than MNC+++, the 1st place entry in 2015. Some example results are visualized in Figure \ref{fig:visualization_coco}.

\paragraph{COCO Detection} The proposed FCIS method also performs well on box-level object detection. By taking the enclosing boxes of the instance masks as detected bounding boxes, it achieves an object detection accuracy of 39.7\% on COCO test-dev set, measured by the standard mAP$^b$@[0.5:0.95] score. The result ranks 2nd in the COCO object detection leaderboard.

\section{Conclusion}
We present the first fully convolutional method for instance-aware semantic segmentation. It extends the existing FCN-based approaches and significantly pushes forward the state-of-the-art in both accuracy and efficiency for the task. The high performance benefits from the highly integrated and efficient network architecture, especially a novel joint formulation.

{\small
\bibliographystyle{ieee}
\bibliography{fcis}
}

\end{document}